\documentclass[runningheads]{llncs}
\usepackage[T1]{fontenc}
\usepackage{graphicx}
\usepackage{amsmath}
\usepackage{amssymb}
\usepackage{booktabs}
\usepackage{xcolor}
\usepackage{multirow}
\usepackage{mathrsfs}
\usepackage{bm}
\usepackage{hyperref}
\usepackage{float}
\usepackage{placeins}

\DeclareMathOperator*{\argmax}{arg\,max}
\newcommand{\info}{\text{info}}
\newcommand{\verificationoperator}{\mathcal{G}}

\definecolor{cvprblue}{rgb}{0.21,0.49,0.74}
\hypersetup{
    colorlinks=true,
    linkcolor=cvprblue,
    filecolor=cvprblue,
    urlcolor=cvprblue,
    citecolor=cvprblue
}

\begin{document}
\title{SARA: Scene-Aware Reconstruction Accelerator}
\titlerunning{SARA}
%
\author{Jee Won Lee\inst{1,2} \
Hansol Lim\inst{1,2} \\
Minhyeok Im\inst{3,4} \
Dohyeon Lee\inst{3,4} \
Jongseong Brad Choi\inst{1,2}\thanks{Corresponding author.}}
\authorrunning{J. W. Lee et al.}
%
\institute{Department of Mechanical Engineering, State University of New York, Korea, Incheon, South Korea 
\and
Department of Mechanical Engineering, State University of New York, Stony Brook, NY, United States 
\and
Department of Computer Science, State University of New York, Korea, Incheon, South Korea 
\and
Department of Computer Science, State University of New York, Stony Brook, NY, United States}
\maketitle              
\begin{abstract}
We present SARA (Scene-Aware Reconstruction Accelerator), a geometry-driven pair selection module for Structure-from-Motion (SfM). Unlike conventional pipelines that select pairs based on visual similarity alone, SARA introduces geometry-first pair selection by scoring reconstruction informativeness---the product of overlap and parallax---before expensive matching. A lightweight pre-matching stage uses mutual nearest neighbors and RANSAC to estimate these cues, then constructs an Information-Weighted Spanning Tree (IWST) augmented with targeted edges for loop closure, long-baseline anchors, and weak-view reinforcement. Compared to exhaustive matching, SARA reduces rotation errors by $46.5 \pm 5.5\%$ and translation errors by $12.5 \pm 6.5\%$ across modern learned detectors, while achieving at most $50\times$ speedup through 98\% pair reduction (from 30,848 to 580 pairs). This reduces matching complexity from quadratic to quasi-linear, maintaining within $\pm 3\%$ of baseline reconstruction metrics for 3D Gaussian Splatting and SVRaster.
\keywords{Structure-from-Motion 
\and Feature Matching 
\and View Graph.}
\end{abstract}
\section{Introduction}
\label{sec:intro}

Learned feature detectors and matchers---SuperPoint, DISK, ALIKED, LightGlue~\cite{detone2018superpoint,tyszkiewicz2020disk,zhao2023aliked,lindenberger2023lightglue}---have become the standard for Structure-from-Motion (SfM) in Novel View Synthesis pipelines~\cite{mildenhall2020nerf,kerbl20233d,barron2022mipnerf360,schonberger2016sfm,Lee2025Micro-splatting,Lim2024LiDAR-3DGS,Lee2025LiteVoxel}. These learned matchers produce rich correspondence sets; however, not all pairs contribute meaningfully to reconstruction. Redundant near-duplicate pairs and low-parallax matches consume computational resources while adding negligible geometric value to bundle adjustment.

Classical SfM pipelines address quadratic pairing through brute-force verification or vocabulary tree pruning. Vocabulary trees organize descriptors hierarchically and query with TF-IDF scoring for sublinear search. This approach has been standard in COLMAP and related systems for over a decade~\cite{nister2006scalable,schonberger2016sfm}. Large-scale reconstruction systems like Photo Tourism~\cite{snavely2006photo}, Building Rome in a Day~\cite{agarwal2009building}, and Building Rome on a Cloudless Day~\cite{frahm2010building} demonstrated scalability through careful view-graph construction and distributed optimization. While efficient, Bag-of-Words scoring over-selects near-duplicates with minimal parallax while missing informative wide-baseline pairs.

Learned retrieval methods improve candidate quality by training global descriptors for place recognition, achieving higher recall than Bag-of-Words. However, these methods share a fundamental limitation: optimizing for visual similarity rather than geometric value for reconstruction.

Recent detector-free methods such as MASt3R and DUSt3R~\cite{leroy2024mast3r,wang2024dust3r} handle extreme geometry well, yet detector-based pipelines remain dominant due to efficiency. The core issue is that similarity-based retrieval identifies co-visible images but cannot predict whether a pair actually improves triangulation conditioning. View selection principles from Multi-View Stereo (MVS)~\cite{furukawa2010accurate,schonberger2016pixelwise}---which balance overlap and parallax---have been largely overlooked in SfM pair selection.

This mismatch leads to inefficient compute allocation. Near-duplicates with negligible parallax consume resources while adding little geometric value. We score pairs by the product of overlap and parallax. Overlap quantifies correspondence availability between views, while parallax determines triangulation conditioning~\cite{hartley1997triangulation}. This multiplicative formulation reflects that both factors are necessary: pairs require sufficient correspondences and adequate baseline separation to improve reconstruction quality.

We present SARA (Scene-Aware Reconstruction Accelerator), which addresses this gap by scoring pairs on \emph{reconstruction informativeness}---defined as overlap $\times$ parallax---before invoking expensive matchers. SARA estimates these geometric cues via lightweight proxies (mutual nearest neighbors + short RANSAC), constructs a maximum-information spanning tree~\cite{kruskal1956shortest}, and augments it with targeted edges for loop closure~\cite{zach2010disambiguating} and weak-view support. The result is fewer matcher calls with better bundle adjustment~\cite{triggs2000bundle} conditioning.

\paragraph{Contributions.}
\begin{itemize}
    \item Geometry-aware pair selection. We introduce a pre-matching scorer that quantifies reconstruction informativeness (overlap $\times$ parallax), constructing an Information-Weighted Spanning Tree (IWST) that maximizes reconstruction quality while minimizing computational cost.
    \item Weak-view reinforcement. We identify and reinforce low-confidence views through targeted edge augmentation, recovering 11.9\% more registered images and preventing reconstruction failure.
    \item Accelerated reconstruction. SARA reduces matching complexity from quadratic to quasi-linear, achieving at best $50\times$ speedup while improving pose accuracy and maintaining within $\pm 3\%$ of baseline metrics~\cite{kerbl20233d} for downstream 3DGS and SVRaster.
\end{itemize}

\begin{figure}[!t]
  \centering
  \includegraphics[width=\columnwidth]{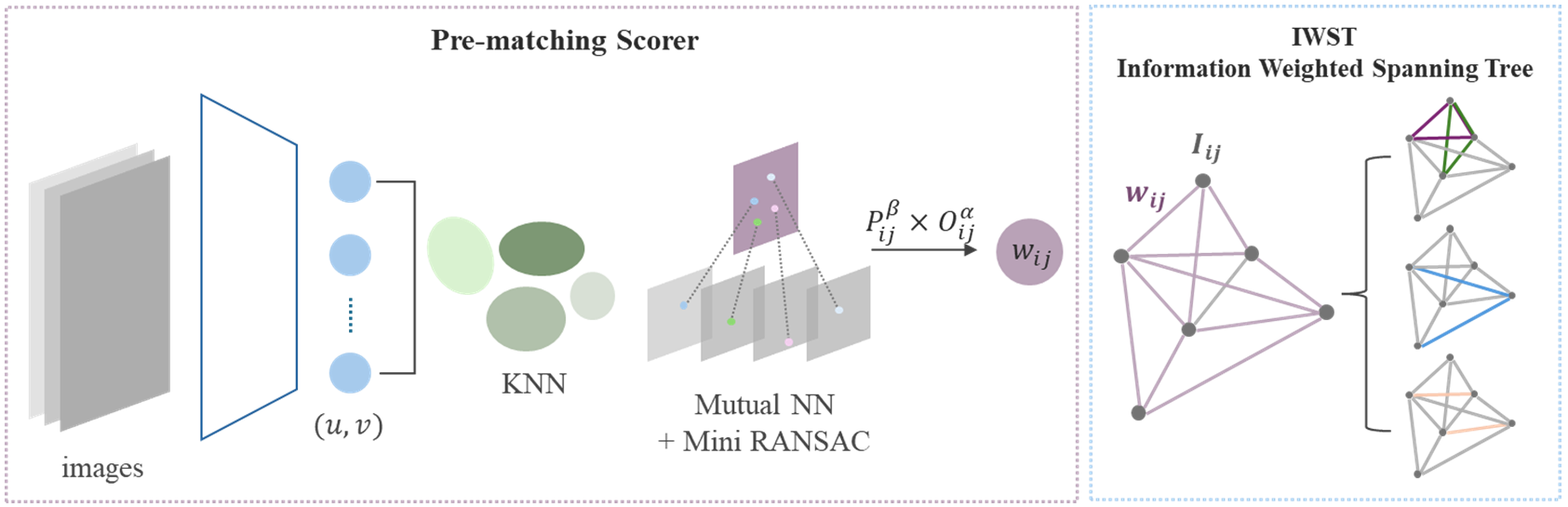}
  \caption{System Overview of SARA: Scene-Aware Reconstruction Accelerator.}
  \label{fig:system_overview}
\end{figure}

\section{Related Work}
\label{sec:related_work}

\paragraph{Feature Extractors.} Classical hand-crafted features like SIFT~\cite{lowe2004distinctive} provided the foundation for SfM pipelines, offering robustness to viewpoint and illumination changes. Learned detectors such as SuperPoint~\cite{detone2018superpoint}, DISK~\cite{tyszkiewicz2020disk}, and ALIKED~\cite{zhao2023aliked} have demonstrated superior repeatability and discriminability through deep learning, achieving state-of-the-art performance on benchmark datasets. However, their improved accuracy comes at the cost of increased computational overhead compared to classical methods.

\paragraph{Image Retrieval and Pair Selection.} Large-scale SfM systems rely on efficient image retrieval to avoid exhaustive pairwise matching. Vocabulary trees~\cite{nister2006scalable} using TF-IDF scoring have been the standard approach in systems like COLMAP~\cite{schonberger2016sfm}, enabling sublinear search through hierarchical descriptor organization. While learned global descriptors have improved retrieval quality through end-to-end training, all these methods optimize for visual similarity rather than geometric informativeness.

\paragraph{Feature Matching.} Classical feature matching relies on nearest-neighbor search with ratio tests and geometric verification through RANSAC~\cite{fischler1981random}. Recent learned matchers like SuperGlue~\cite{sarlin2020superglue}, LightGlue~\cite{lindenberger2023lightglue}, and detector-free approaches such as LoFTR~\cite{sun2021loftr} and DUSt3R~\cite{wang2024dust3r} achieve remarkable accuracy through attention mechanisms and dense correspondence prediction. Despite these advances, computational cost remains a bottleneck, making efficient pair selection critical for practical reconstruction systems.

\paragraph{} SARA addresses this gap by introducing geometry-aware pair selection that optimizes for reconstruction informativeness rather than visual similarity. Unlike existing methods, SARA uses lightweight geometric proxies to estimate overlap and parallax before expensive matching, reducing computational cost while improving reconstruction quality.
\section{Preliminaries}
\label{sec:preliminaries}

\subsection{Generalized Feature Detector.}
Let $I: \Omega \subset \mathbb{R}^2 \to \mathbb{R}^c$ be an image. A generalized feature detector is a parametrized map
\begin{equation}
    D_{\theta}: I \mapsto \{x_i, \sigma_i, \alpha_i, s_i, f_i\}_{i=1}^N,
\end{equation}
where each detection, $x_i \in \Omega$ is keypoint location, $\sigma_i > 0$ is scale, $\alpha_i \in (-\pi, \pi]$ is orientation, $s_i \in \mathbb{R}$ is detection confidence (score), and $f_i \in \mathbb{R}^d$ is a unit-normalized descriptor. This covers classical hand-crafted detectors (descriptors) such as SIFT, ORB, and learned detectors such as SuperPoint, ALIKED, DISK~\cite{lowe2004distinctive,rublee2011orb,detone2018superpoint,zhao2023aliked,tyszkiewicz2020disk}. We also subsume detector-free frontends like LoFTR~\cite{sun2021loftr} by viewing $D_{\theta}$ as sampling a dense grid $\{x\}$ with per-pixel features $f(x)$ and optional saliency $s(x)$, in this case $\sigma_i, \alpha_i$ may be omitted and $N=|\Omega|$ or a pruned subset.

Given two images $I_a, I_b$, we denote their detections as
\begin{align}
    D_{\theta}(I_a) &= \{x_{i_a}, \sigma_{i_a}, \alpha_{i_a}, s_{i_a}, f_{i_a}\}_{i=1}^{N_a}, \\
    D_{\theta}(I_b) &= \{x_{j_b}, \sigma_{j_b}, \alpha_{j_b}, s_{j_b}, f_{j_b}\}_{j=1}^{N_b}.
\end{align}
We write $T_{det}(\theta, I)$ for detection time and $N(\theta, I)$ for emitted keypoint count. Both vary with architecture and image content. Memory scales as $O(Nd)$.

\subsection{Generalized Feature Matcher.}
A generalized matcher takes the detector outputs or raw images in detector-free mode and produces a tentative correspondence with confidences:
\begin{equation}
    M_{\phi}: (D_{\theta}(I_a), D_{\theta}(I_b)) \mapsto \mathcal{C}_{ab} = \{(i, j), \pi_{ij}\},
\end{equation}
where $(i, j)$ indexes a putative match between $x_{i_a}$ and $x_{j_b}$, and $\pi_{ij} \in [0, 1]$ is a match confidence (score). This covers both nearest-neighbor matchers, learned matchers, and detector-free correlators.

We define a verification operator $\verificationoperator$ that enforces epipolar consistency and yields inliers and optionally relative pose:
\begin{equation}
    \verificationoperator(\mathcal{C}_{ab}) \Rightarrow \mathcal{I}_{ab}, \mathbf{E}_{ab}, \quad \mathcal{I}_{ab} \subseteq \mathcal{C}_{ab},
\end{equation}
where $\mathbf{E}_{ab}$ is estimated via a robust minimal solver (5-point for calibrated~\cite{nister2004efficient}, 7 or 8 point otherwise~\cite{hartley1997defense}) with RANSAC~\cite{fischler1981random} (or other -SAC family algorithms), and $\mathcal{I}_{ab}$ is the inlier count. If intrinsics are known, $\mathbf{E}_{ab}$ recovers rotation and translation up to scale; otherwise the fundamental matrix $\mathbf{F}_{ab}$ is used~\cite{hartley2003multiple}.

Let $N_a, N_b$ be keypoint counts and $k$ the number of descriptor comparisons per keypoint after candidate pruning. A generic cost decomposition is
\begin{equation}
    T_{match} \approx T_{cand} + O(k N_a) + T_{verify}(|\mathcal{C}_{ab}|),
\end{equation}
where $T_{cand}$ accounts for Approximated Nearest Neighbor look up or attention passes in learned matchers, and $T_{verify}$ scales with the number of hypotheses and sampled subsets.

\paragraph{Geometric Informativeness.}
Pair overlap and parallax. Given verified inliers $\mathcal{I}_{ab}$, we approximate overlap as $O_{ab} = \frac{|\mathcal{I}_{ab}|}{\sqrt{N_a N_b}}$, and parallax using the median triangulation angle $\angle_{ab}$ from $\mathbf{E}_{ab}$ and $\mathcal{I}_{ab}$~\cite{hartley1997triangulation}. We define Informativeness by scoring a candidate pair by
\begin{equation}
    \info_{ab} \propto O_{ab} \times g(\angle_{ab}),
\end{equation}
with $g(\cdot)$ monotone on a practical range (e.g., $g(x) = \min(x, C)$ or $g(x) = \tanh(x/\tau)$ for some constant $C$ or $\tau$, saturating beyond wide baselines to prevent instability). This coupling reflects fundamental reconstruction principles: overlap ensures matchability while parallax controls triangulation uncertainty~\cite{hartley2003multiple}. Narrow baselines amplify depth errors quadratically~\cite{scharstein2002taxonomy,hartley1997triangulation}, while insufficient overlap prevents matching. Prior work in multi-view stereo has demonstrated that view selection based on balancing overlap and parallax substantially improves reconstruction quality~\cite{furukawa2010accurate,schonberger2016pixelwise,gallup2008variable}, yet similarity-based retrieval ignores this geometric trade-off. Given a candidate set $\mathcal{R}_a \subset \{b\}$ from any retrieval method, SARA re-ranks $\mathcal{R}_a$ using $\info_{ab}$ before invoking $M_{\phi}$.
\section{Methodology}
\label{sec:methodology}

\subsection{Overview}
SARA builds a two-layer view graph to achieve quasi linear matching complexity while preserving reconstruction quality. It first uses a lightweight pre-matching scorer to assign a geometric informativeness weight, $w_{ij}$, to each candidate pair. It then constructs an Information-Weighted Spanning Tree (IWST) as a sparse, geometrically-guided backbone, which is augmented with a small set of edges to ensure robustness.

\subsection{Pre-matching Scorer}
We score each candidate pair before any expensive matching, using a two-step procedure where we first prune to top-$k$ neighbors with global descriptors then extract a tiny, reliable local sample via Mutual-NN and validate it with a short RANSAC. The result is a single informativeness weight $w_{ij}$ used as edge weights.

\begin{figure}[!t]
  \centering
  \includegraphics[width=\linewidth]{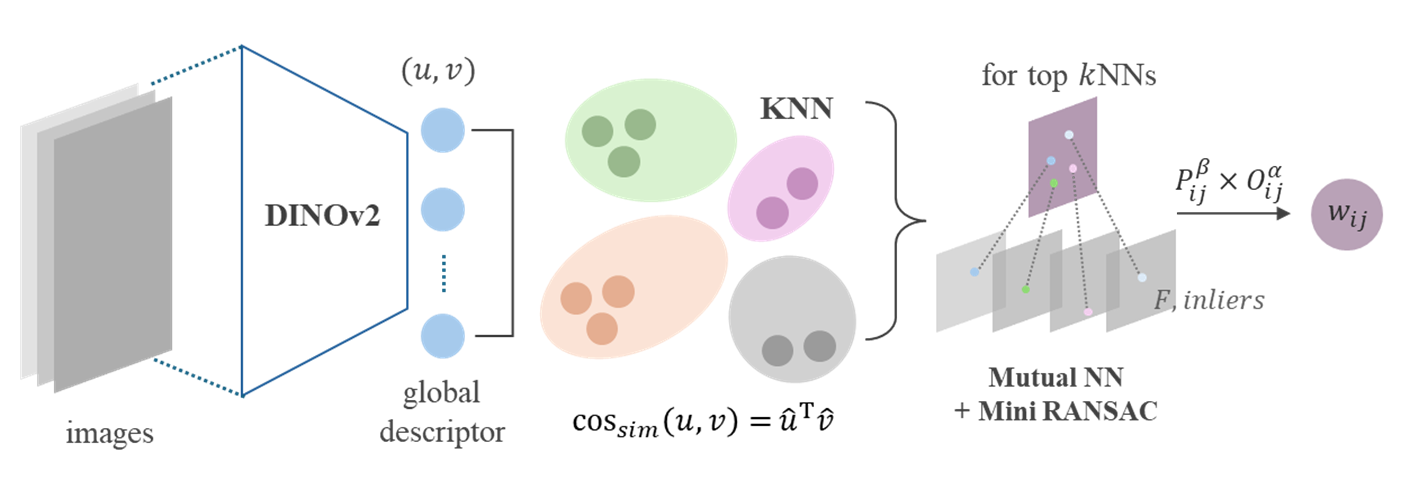}
  \caption{Pre-matching scorer.}
  \label{fig:prematching_scorer}
\end{figure}

\paragraph{Candidate Pruning.}
For each image $i$, we compute one global descriptor per image with DINOv2~\cite{oquab2023dinov2}. Let $\mathbf{u}_i \in \mathbb{R}^d$ be the L2-normalized descriptor for image $i$. We rank neighbors $j$ by cosine similarity
\begin{equation}
    \mathrm{sim}(\mathbf{u}_i, \mathbf{u}_j) = \mathbf{u}_i^T \mathbf{u}_j,
\end{equation}
and keep the top-$k$ per image. The retrieval set is the symmetric union
\begin{equation}
    \mathcal{R} = \{(i, j) | j \in \text{Top-}k(i) \text{ or } i \in \text{Top-}k(j)\}
\end{equation}
This reduces the quadratic search to a small $\mathcal{R}$ while preserving likely overlaps.

For each candidate pair $(i, j)$ from kNN retrieval, we gather the top-$b$ mutual descriptor matches ($\mathcal{C}_{ij}$, e.g., $b=50$). Mutual nearest neighbors provide a high-confidence subset, reducing outlier ratio compared to one-way matching. We apply standard RANSAC~\cite{fischler1981random} with reduced scope---32 iterations on $\leq b$ correspondences versus hundreds of iterations on thousands---which we term short RANSAC. This is sufficient for scoring pair quality rather than final pose estimation, as we only need coarse geometric signals to rank pairs.

If the intrinsics are known, we estimate the essential matrix $\mathbf{E}_{ij}$ with a 5-point solver~\cite{nister2004efficient} and otherwise the fundamental matrix $\mathbf{F}_{ij}$ via 7/8-point~\cite{hartley1997defense}. We score each hypothesis by inlier count under the Sampson error~\cite{hartley2003multiple}, a closed-form first-order approximation to reprojection error. Unlike reprojection error, which requires iterative triangulation for each correspondence, Sampson error evaluates geometric consistency directly from the epipolar constraint, providing efficient verification suitable for pre-matching. We retain the best inlier set $\mathcal{I}_{ij}$. From $\mathbf{E}_{ij}$ we recover $\mathbf{R}, \mathbf{t}$ and triangulate~\cite{hartley1997triangulation} $\mathcal{I}_{ij}$ to obtain two coarse signals,
\begin{align}
    O_{ij} &= \frac{|\mathcal{I}_{ij}|}{\sqrt{N_i N_j}}, \\ P_{ij} &= \mathrm{median}_{k \in \mathcal{I}_{ij}} (\theta_k),
\end{align}
where $O_{ij}$ is the overlap normalized by $\sqrt{N_i N_j}$, and $P_{ij}$ is the parallax measured as the median bearing angle. The median operator is robust to outliers in the limited correspondence set. The pre-matching informativeness then becomes
\begin{equation}
    w_{ij} = O_{ij}^{\alpha} P_{ij}^{\beta},
\end{equation}
with rejection if $O_{ij} < \tau_O$ or $P_{ij} < \tau_P$. This design preserves the ability to discriminate useful pairs while making the computation markedly cheaper than full matching and full RANSAC, enabling scoring over $\mathcal{R}$ candidates within the pre-matching budget.

\subsection{Information Weighted Spanning Tree (IWST)}
Let $G=(V, E, w)$ be the view-graph whose edges carry the pre-matching informativeness weights, $w_{ij}$ from \S b. We first build a thin, connected backbone by maximizing the total weight over spanning trees~\cite{kruskal1956shortest,prim1957shortest}:
\begin{equation}
    T^* = \argmax_{T \in \mathcal{ST}(G)} \sum_{(i,j) \in T} w_{ij}.
\end{equation}
After $T^*$, we add a small, orthogonal set of non-tree edges to address failure modes not optimized by the tree objective. Let $\mathcal{C} = E \setminus T^*$ denote candidate augmentations and let $B$ be a tiny edge budget where we score $(i,j) \in \mathcal{C}$ by role and greedily add the best few, yielding $T_{aug}^*$.

We then augment $T^*$ with three targeted additions—multi-scale loop consolidation, long-baseline anchors, and weak-view reinforcement—that respectively distribute drift, stabilize global scale, and support low-confidence views before proceeding to full matching as shown in Fig. 3.

\paragraph{Multi-Scale Loop Consolidation (WSL).}
For loop closures, we prefer edges that close short/medium/long cycles on $T^*$~\cite{zach2010disambiguating,murartal2017orbslam2}. For an edge $(i, j)$, let $\text{path}_{T^*}(i,j)$ denote the unique tree path. We define a loop score
\begin{equation}
    s_{ij}^{\text{loop}} = w_{ij} \cdot \gamma(l_{ij}), \quad l_{ij} = |\text{path}_{T^*}(i,j)|,
\end{equation}
where $\gamma(\cdot)$ encourages a mix of short/ medium/long cycles, thereby adding up to $B_{\text{loop}}$ such as edges distributes pose error and damps drift across the scaffold.

\begin{figure}[!t]
  \centering
  \includegraphics[width=0.8\linewidth]{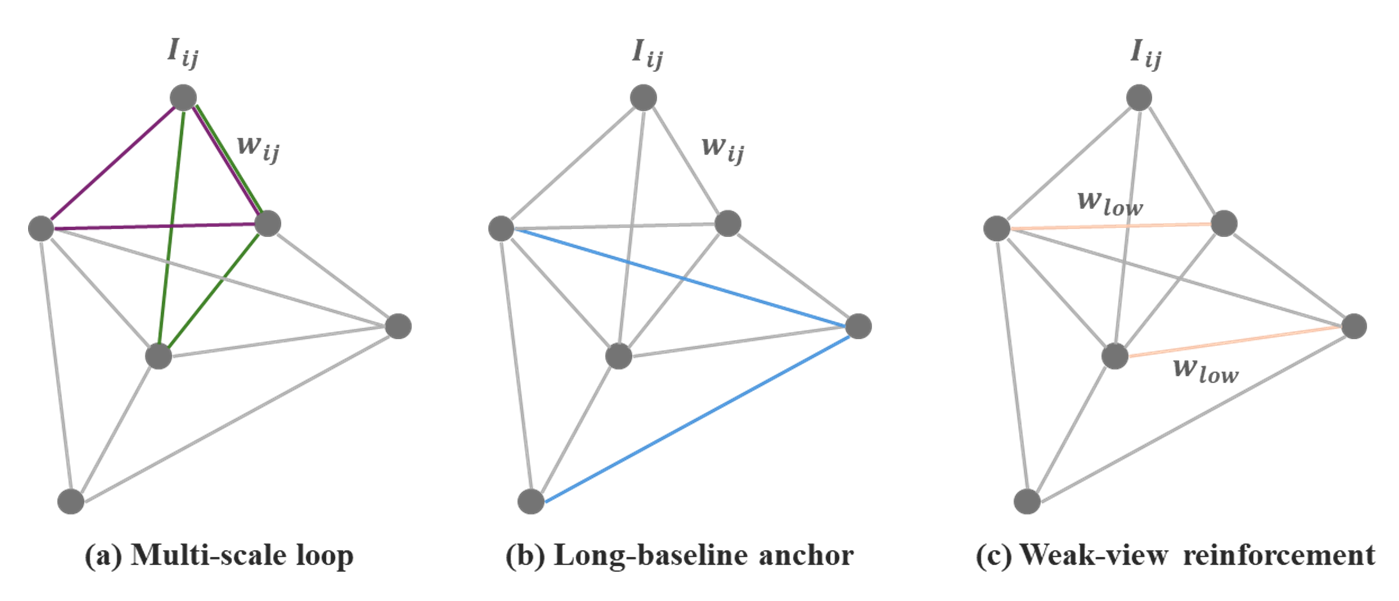}
  \caption{Backbone augmentation after IWST.}
  \label{fig:backbone_augmentation}
\end{figure}

\paragraph{Long-Baseline Anchors (LBA).}
To stabilize global scale and improve triangulation conditioning, we prioritize large-parallax pairs using the same $P_{ij}$ from \S b and score
\begin{equation}
    s_{ij}^{\text{anchor}} = P_{ij} \cdot w_{ij},
\end{equation}
adding up to $B_{\text{anc}}$ diverse anchors once loops have stabilized the graph.

\paragraph{Weak-View Reinforcement (WVR).}
Let $\deg_{T^*}(i)$ be the node degree in $T^*$ and $\kappa_i$ a per-view confidence proxy, the median $w_{ij}$ on incident edges. For candidates touching a weak view $u$ (low degree or $\kappa$), we score
\begin{equation}
    s_{uj}^{\text{weak}} = w_{uj} \cdot \rho(\deg_{T^*}(u), \kappa_u),
\end{equation}
and add up to $B_{\text{weak}}$ such edges so low-weight views remain connected and recoverable.

We add multi-scale loops first to redistribute and damp drift on the whole tree, long-baseline anchors second to lock global scale on the stabilized graph, and weak-view reinforcement last so low-weight views gain support without propagating error, achieving maximal reliability with minimal extra edges.

Consequently, the backbone construction uses the same informative weights learned in \S b to (i) produce an $O(|V|)$-edge core optimized for geometry and (ii) attach a few targeted augmentations under a tiny budget $B = B_{\text{loop}} + B_{\text{anc}} + B_{\text{weak}}$. This delivers a connected and well-conditioned view-graph while reducing matcher calls from quadratic to quasi-linear complexity.

\begin{center}
\small
\begin{tabular}{@{}p{0.95\columnwidth}@{}}
\toprule
Algorithm 1 Pre-Matching Scorer \\
\midrule
Procedure \textsc{PreMatchingScorer}($\mathcal{I}, D_{\theta}, k$) \\
\quad Input: Image set $\mathcal{I} = \{I_1, \dots, I_N\}$, detector $D_{\theta}$, \\
\quad \quad \quad kNN budget $k$ \\
\quad Output: Edge weights $\{w_{ij}\}$ \\
1: Extract global descriptors: $\mathbf{u}_i \leftarrow \text{GlobalDesc}(I_i)$ $\forall i$ \\
2: Retrieve candidates: $\mathcal{R} \leftarrow \text{kNN}(\{\mathbf{u}_i\}, k)$ \\
3: for each pair $(i,j) \in \mathcal{R}$ do \\
4: \quad Extract top-$b$ mutual-NN matches $\mathcal{C}_{ij}$ \\
\quad \quad \quad between $D_{\theta}(I_i), D_{\theta}(I_j)$ \\
5: \quad Run short RANSAC on $\mathcal{C}_{ij}$ to get $\mathbf{E}_{ij}$, inliers $\mathcal{I}_{ij}$ \\
6: \quad Compute overlap: $O_{ij} = |\mathcal{I}_{ij}| / \sqrt{N_i N_j}$ \\
7: \quad Compute parallax: $P_{ij} = \text{median}_{k \in \mathcal{I}_{ij}}(\theta_k)$ \\
8: \quad Compute informativeness: $w_{ij} = O_{ij}^{\alpha} \cdot P_{ij}^{\beta}$ \\
9: \quad if $O_{ij} < \tau_O$ or $P_{ij} < \tau_P$ then reject pair $(i, j)$ \\
10: end for \\
11: return $\{w_{ij}\}$ \\
End Procedure \\
\bottomrule
\end{tabular}
\end{center}

\begin{center}
\small
\begin{tabular}{@{}p{0.95\columnwidth}@{}}
\toprule
Algorithm 2 Two-Layer View-Graph Construction \\
\midrule
Procedure \textsc{ViewGraphConstruction}($\mathcal{R}, \{w_{ij}\}$) \\
\quad Input: Candidate pairs $\mathcal{R}$, edge weights $\{w_{ij}\}$ \\
\quad Output: View-graph edges $E$ for full matching \\
1: Compute maximum spanning tree: \\
\quad $T^* = \argmax_{T \in \mathcal{ST}(G)} \sum_{(i,j) \in T} w_{ij}$ \\
2: Initialize $E \leftarrow T^*$, $\mathcal{C} \leftarrow \mathcal{R} \setminus T^*$ \\
3: Add multi-scale loops: $E \leftarrow E \cup \text{TopK}(\mathcal{C}, s_{ij}^{\text{loop}}, B_{\text{loop}})$ \\
4: Add long-baseline anchors: \\
\quad $E \leftarrow E \cup \text{TopK}(\mathcal{C} \setminus E, s_{ij}^{\text{anchor}}, B_{\text{anc}})$ \\
5: Add weak-view reinforcement: \\
\quad $E \leftarrow E \cup \text{TopK}(\mathcal{C} \setminus E, s_{ij}^{\text{weak}}, B_{\text{weak}})$ \\
6: return $E$ \\
End Procedure \\
\bottomrule
\end{tabular}
\end{center}
\section{Results and Evaluation}
\label{sec:results}

\subsection{Quantitative Analysis}

We evaluate SARA on the \textit{Mip-NeRF 360} dataset~\cite{barron2022mipnerf360}, which consists of 5 challenging indoor and outdoor scenes (\textit{bonsai}, \textit{garden}, \textit{kitchen}, \textit{room}, \textit{stump}). 

\begin{table}[!htbp]
\centering
\caption{Comprehensive performance comparison. SARA achieves 100\% image registration with over $40\times$ speedup in matching time.}
\label{tab:main_results}
\resizebox{\columnwidth}{!}{
\begin{tabular}{@{}llrrrrrrrrrr@{}}
\toprule
Method & Extractor & Reg. & Rot. & Trans. & Matches & Mean & Median & 3D & Track & Pairs & Match \\
& & (\%) & Error & Error & /Pair & Reproj. & Reproj. & Points & Length & & Time (s) \\ \midrule
Exhaustive & SUPERPOINT & 89.5 & 45.9 & 40.8 & 484 & 1.20 & 1.19 & 74003 & 10.7 & 30848 & 734 \\
 & ALIKED & 81.7 & 40.4 & 37.6 & 539 & 0.93 & 0.84 & 65269 & 13.7 & 30848 & 692 \\
 & DISK & 78.5 & 34.5 & 36.5 & 531 & 0.89 & 0.77 & 69751 & 12.3 & 30848 & 669 \\ \midrule
VocabTree & SUPERPOINT & 92.4 & 39.3 & 36.0 & 483 & 0.79 & 0.81 & 54612 & 8.2 & 2274 & 54 \\
 & ALIKED & 86.6 & 37.0 & 30.9 & 536 & 0.62 & 0.60 & 63456 & 11.4 & 2805 & 66 \\
 & DISK & 83.8 & 32.8 & 28.4 & 596 & 0.65 & 0.63 & 55331 & 9.1 & 2933 & 70 \\ \midrule
SARA & SUPERPOINT & 100.0 & 21.4 & 33.1 & 1083 & 0.75 & 0.76 & 54089 & 6.8 & 572 & 14 \\
 & ALIKED & 100.0 & 22.0 & 32.3 & 1235 & 0.58 & 0.53 & 48160 & 9.4 & 589 & 14 \\
 & DISK & 100.0 & 20.2 & 34.4 & 1206 & 0.58 & 0.54 & 42086 & 8.2 & 571 & 14 \\ \bottomrule
\end{tabular}
}
\end{table}

We compare SARA against two standard baselines: Exhaustive Matching, the quadratic all-pairs matching approach, and Vocabulary Tree~\cite{nister2006scalable}, a widely-used visual similarity retrieval method configured with k=30 nearest neighbors. To demonstrate SARA's extractor-agnostic design, we test all methods using three modern learned feature extractors: SuperPoint, ALIKED, and DISK~\cite{detone2018superpoint,zhao2023aliked,tyszkiewicz2020disk}.

We assess performance across three main categories: (1) Pose Estimation Quality (rotation error, translation error, image registration rate), (2) Reconstruction Quality (reprojection error, 3D points, track length), and (3) Computational Efficiency (number of pairs matched, matching time). We explicitly do not report pose estimation AUC, as this metric primarily evaluates the robustness of the pose estimation algorithm (e.g., RANSAC) rather than the quality of the pair selection method itself.

We excluded SIFT from our main analysis due to incompatibility with SARA's design assumptions. SIFT produces noisy correspondences on the \textit{Mip-NeRF 360} dataset, failing RANSAC geometric validation in our Pre-Matching Scorer. This results in sparse view-graphs with insufficient connectivity for reconstruction. In contrast, exhaustive matching can register views despite low-quality features by brute-forcing all pairs. This outcome reflects SARA's design: it is a precision-oriented filter that requires reliable input features, not a corrective mechanism for low-quality detectors.

As shown in Table \ref{tab:main_results}, SARA's geometry-aware selection achieves 100\% image registration across all modern detectors, outperforming both baselines. The key insight is that overlap $\times$ parallax captures the two fundamental requirements for accurate triangulation: (1) \emph{overlap} ensures sufficient correspondences exist between views, and (2) \emph{parallax} determines triangulation precision, since depth uncertainty scales inversely with baseline angle~\cite{hartley1997triangulation}. Neither factor alone suffices---high overlap with zero parallax yields no depth information, while high parallax with no overlap yields no matches. SARA's multiplicative scoring naturally balances these competing demands.

SARA’s performance is extractor-agnostic. The pose and quality improvements are consistent across all modern learned detectors (SuperPoint, ALIKED, and DISK). This confirms SARA’s robust, geometry-driven design. In contrast, vocabulary tree’s performance varies significantly, as its appearance-based retrieval is highly dependent on descriptor quality.

\FloatBarrier
\subsection{Application to Novel View Synthesis}

Many state-of-the-art novel view synthesis (NVS) pipelines depend on Structure-from-Motion (SfM) priors and accurate camera poses. To verify that SARA's geometry-aware pair selection is not only computationally efficient but also downstream-safe, we evaluate its impact on two representative SfM-prior-based renderers, 3D Gaussian Splatting (3DGS)~\cite{kerbl20233d} and SVRaster, on the Mip-NeRF 360 dataset~\cite{barron2022mipnerf360} over four scenes (bonsai, garden, stump, room), with all experiments executed on a single RTX 5090 GPU.

We build COLMAP reconstructions using three detector--matcher backbones (ALIKED + LightGlue, DISK + LightGlue, and SuperPoint + LightGlue) and three match-selection strategies (exhaustive all-pairs, vocabulary-tree retrieval, and SARA), yielding nine SfM configurations (a--i): (a) ALIKED + LightGlue with exhaustive matching, (b) ALIKED + LightGlue with vocabulary tree, (c) ALIKED + LightGlue with SARA, (d) DISK + LightGlue with exhaustive matching, (e) DISK + LightGlue with vocabulary tree, (f) DISK + LightGlue with SARA, (g) SuperPoint + LightGlue with exhaustive matching, (h) SuperPoint + LightGlue with vocabulary tree, and (i) SuperPoint + LightGlue with SARA. For each configuration, we train 3DGS and SVRaster using the resulting poses while keeping renderer-side settings fixed, isolating the effect of pair selection on NVS quality. In the qualitative figures, the rows labeled ``3DGS'' and ``SVRaster'' correspond to the original renderer outputs produced from the standard COLMAP SfM pipeline (SIFT features with RANSAC-based verification), serving as a conventional SfM prior baseline for each renderer.

\begin{figure}[H]
  \centering
  \includegraphics[width=0.9\linewidth]{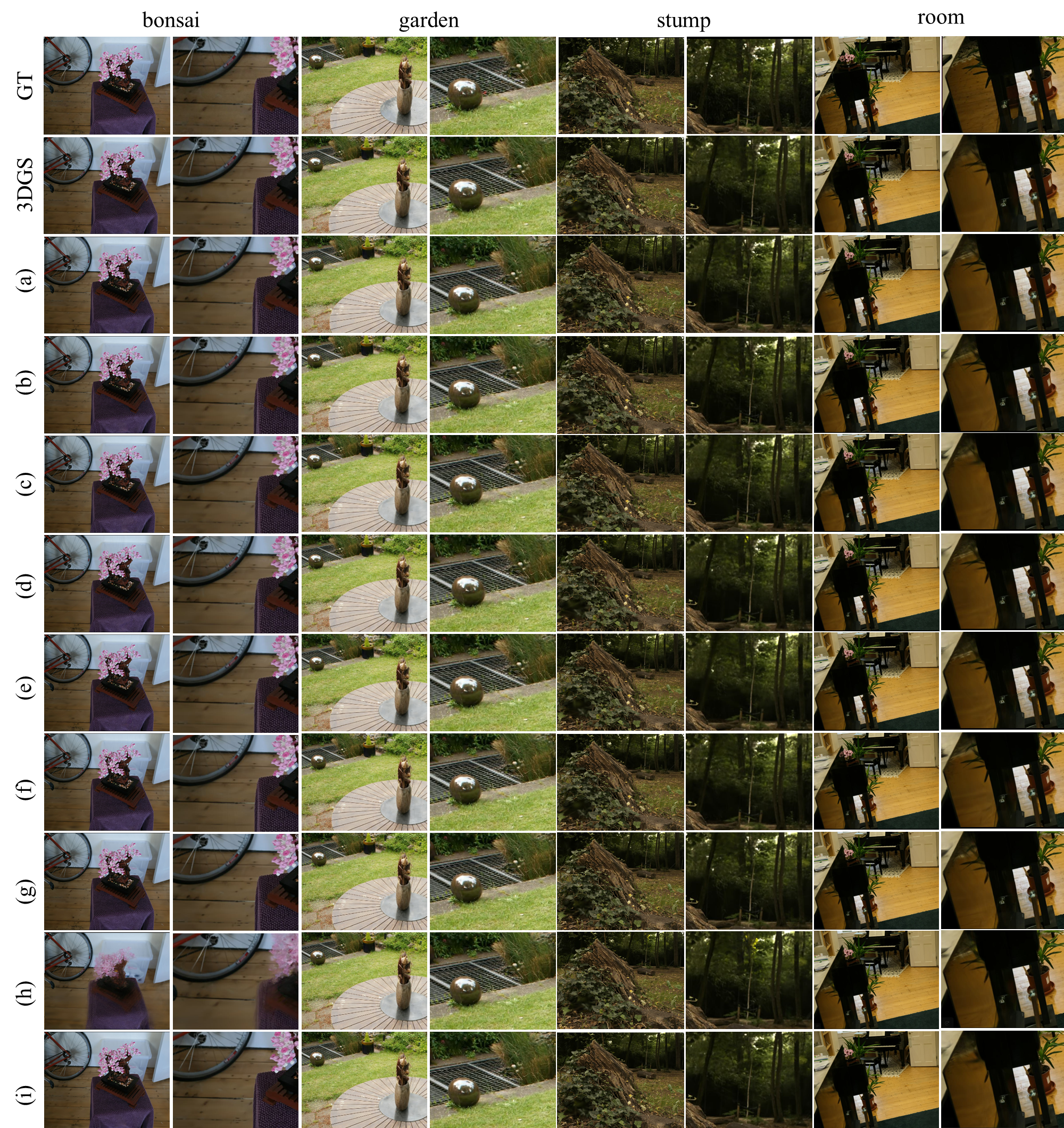}
  \caption{Qualitative comparison of novel view synthesis results using 3D Gaussian Splatting across different SfM configurations.}
  \label{fig:3dgs_qualitative}
\end{figure}

\paragraph{Results on 3DGS.} Across scenes, SARA-based SfM priors (c,f,i) yield novel-view renderings that are visually indistinguishable from exhaustive matching (a,d,g), without introducing systematic blur, ghosting, or geometric instability where representative comparisons are shown in Fig.~\ref{fig:3dgs_qualitative}. This qualitative consistency is aligned by the quantitative metrics (SSIM/PSNR/LPIPS and L1), where SARA tracks the exhaustive baseline closely across detector families, as summarized in Table~\ref{tab:3dgs_results}. In contrast, vocabulary-tree selection exhibits higher variance and can underperform, indicating that appearance-only retrieval is less reliable at preserving the geometrically informative constraints needed for stable SfM priors and downstream splatting quality.

\begin{table}[H]
\centering
\caption{Quantitative evaluation of 3DGS rendering quality across different SfM configurations. SARA-based reconstructions (c,f,i) achieve comparable quality to exhaustive matching (a,d,g).}
\label{tab:3dgs_results}
\resizebox{\columnwidth}{!}{
\begin{tabular}{@{}l|cccc|cccc|cccc|cccc@{}}
\toprule
& \multicolumn{4}{c|}{bonsai} & \multicolumn{4}{c|}{garden} & \multicolumn{4}{c|}{stump} & \multicolumn{4}{c}{room} \\
Method & SSIM$\uparrow$ & L1$\downarrow$ & PSNR$\uparrow$ & LPIPS$\downarrow$ & SSIM$\uparrow$ & L1$\downarrow$ & PSNR$\uparrow$ & LPIPS$\downarrow$ & SSIM$\uparrow$ & L1$\downarrow$ & PSNR$\uparrow$ & LPIPS$\downarrow$ & SSIM$\uparrow$ & L1$\downarrow$ & PSNR$\uparrow$ & LPIPS$\downarrow$ \\ \midrule
3DGS & 0.967 & 0.010 & 36.52 & 0.053 & 0.923 & 0.018 & 31.57 & 0.052 & 0.899 & 0.021 & 28.38 & 0.096 & 0.937 & 0.014 & 32.93 & 0.073 \\
(a) & 0.953 & 0.011 & 35.11 & 0.061 & 0.913 & 0.019 & 31.27 & 0.061 & 0.888 & 0.023 & 26.98 & 0.108 & 0.936 & 0.014 & 32.63 & 0.078 \\
(b) & 0.954 & 0.011 & 35.11 & 0.061 & 0.915 & 0.019 & 31.35 & 0.059 & 0.893 & 0.022 & 27.47 & 0.100 & 0.936 & 0.014 & 33.08 & 0.075 \\
(c) & 0.952 & 0.012 & 35.00 & 0.061 & 0.906 & 0.020 & 30.94 & 0.066 & 0.897 & 0.021 & 28.03 & 0.098 & 0.930 & 0.015 & 31.96 & 0.091 \\
(d) & 0.955 & 0.011 & 35.46 & 0.057 & 0.916 & 0.019 & 31.24 & 0.061 & 0.888 & 0.022 & 28.05 & 0.108 & 0.938 & 0.013 & 33.28 & 0.071 \\
(e) & 0.956 & 0.011 & 35.41 & 0.057 & 0.915 & 0.019 & 31.14 & 0.062 & 0.883 & 0.022 & 27.89 & 0.106 & 0.939 & 0.014 & 33.16 & 0.071 \\
(f) & 0.951 & 0.012 & 33.75 & 0.067 & 0.911 & 0.019 & 31.05 & 0.066 & 0.889 & 0.019 & 30.78 & 0.098 & 0.944 & 0.013 & 32.52 & 0.057 \\
(g) & 0.957 & 0.011 & 35.64 & 0.055 & 0.918 & 0.019 & 31.46 & 0.058 & 0.885 & 0.023 & 27.06 & 0.111 & 0.939 & 0.014 & 33.20 & 0.068 \\
(h) & 0.826 & 0.029 & 23.74 & 0.207 & 0.917 & 0.019 & 31.35 & 0.059 & 0.882 & 0.023 & 27.39 & 0.111 & 0.940 & 0.014 & 33.29 & 0.069 \\
(i) & 0.957 & 0.011 & 35.50 & 0.056 & 0.914 & 0.019 & 31.31 & 0.064 & 0.907 & 0.018 & 31.41 & 0.064 & 0.935 & 0.014 & 32.91 & 0.075 \\ \bottomrule
\end{tabular}
}
\end{table}

\begin{figure}[H]
  \centering
  \includegraphics[width=0.9\linewidth]{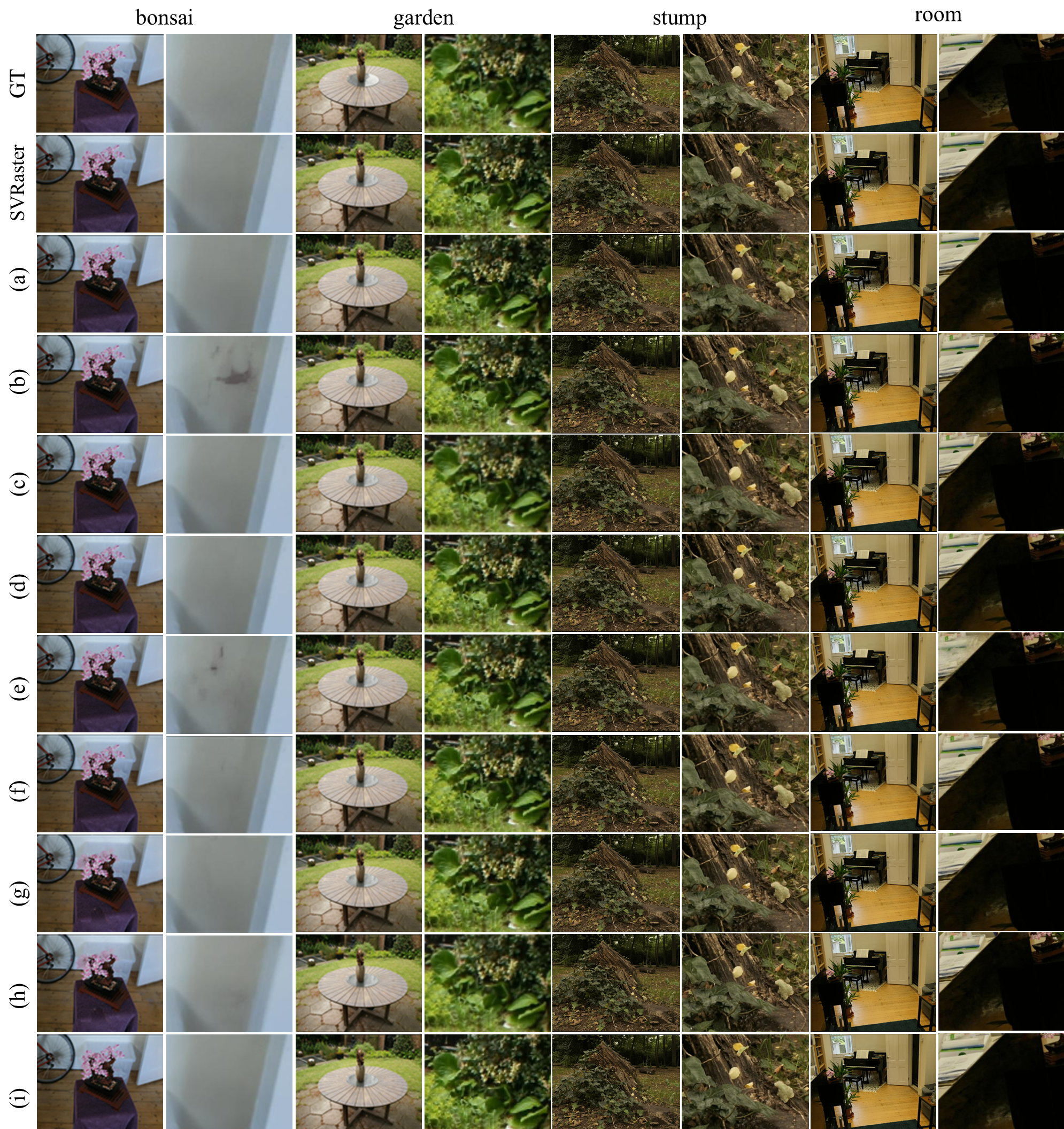}
  \caption{Qualitative comparison of novel view synthesis results using SVRaster across different SfM configurations.}
  \label{fig:svraster_qualitative}
\end{figure}

\paragraph{Results on SVRaster.} The same trend holds when swapping the renderer to SVRaster: SARA-based reconstructions maintain parity with exhaustive matching in both perceptual quality and metric fidelity, as shown in Fig.~\ref{fig:svraster_qualitative} and Table~\ref{tab:svraster_results}. This is significant because it demonstrates that SARA's pair selection does not overfit to a specific renderer's optimization dynamics. Instead, it preserves the integrity of SfM priors in a way that generalizes across two materially different NVS pipelines.

\begin{table}[H]
\centering
\caption{Quantitative evaluation of SVRaster rendering quality across different SfM configurations. SARA-based reconstructions maintain competitive quality with exhaustive matching.}
\label{tab:svraster_results}
\resizebox{\columnwidth}{!}{
\begin{tabular}{@{}l|cccc|cccc|cccc|cccc@{}}
\toprule
& \multicolumn{4}{c|}{bonsai} & \multicolumn{4}{c|}{garden} & \multicolumn{4}{c|}{stump} & \multicolumn{4}{c}{room} \\
Method & SSIM$\uparrow$ & L1$\downarrow$ & PSNR$\uparrow$ & LPIPS$\downarrow$ & SSIM$\uparrow$ & L1$\downarrow$ & PSNR$\uparrow$ & LPIPS$\downarrow$ & SSIM$\uparrow$ & L1$\downarrow$ & PSNR$\uparrow$ & LPIPS$\downarrow$ & SSIM$\uparrow$ & L1$\downarrow$ & PSNR$\uparrow$ & LPIPS$\downarrow$ \\ \midrule
SVRaster & 0.962 & 0.014 & 33.38 & 0.161 & 0.888 & 0.023 & 29.46 & 0.145 & 0.915 & 0.018 & 31.78 & 0.206 & 0.954 & 0.014 & 33.71 & 0.179 \\
(a) & 0.954 & 0.015 & 32.53 & 0.166 & 0.883 & 0.023 & 29.30 & 0.157 & 0.910 & 0.018 & 31.47 & 0.211 & 0.952 & 0.014 & 33.50 & 0.182 \\
(b) & 0.955 & 0.016 & 32.65 & 0.166 & 0.883 & 0.023 & 29.32 & 0.158 & 0.911 & 0.018 & 31.52 & 0.209 & 0.952 & 0.014 & 33.49 & 0.183 \\
(c) & 0.952 & 0.016 & 32.35 & 0.169 & 0.879 & 0.023 & 29.21 & 0.156 & 0.909 & 0.019 & 31.42 & 0.211 & 0.951 & 0.015 & 33.40 & 0.182 \\
(d) & 0.956 & 0.015 & 32.79 & 0.165 & 0.879 & 0.024 & 28.97 & 0.160 & 0.912 & 0.018 & 31.62 & 0.210 & 0.953 & 0.014 & 33.58 & 0.181 \\
(e) & 0.957 & 0.015 & 32.91 & 0.164 & 0.878 & 0.023 & 28.93 & 0.159 & 0.908 & 0.019 & 31.36 & 0.212 & 0.952 & 0.014 & 33.52 & 0.182 \\
(f) & 0.934 & 0.018 & 31.05 & 0.202 & 0.877 & 0.024 & 28.92 & 0.158 & 0.914 & 0.018 & 31.69 & 0.202 & 0.951 & 0.014 & 33.53 & 0.186 \\
(g) & 0.957 & 0.015 & 32.84 & 0.164 & 0.883 & 0.023 & 29.47 & 0.155 & 0.912 & 0.018 & 31.62 & 0.211 & 0.953 & 0.014 & 33.57 & 0.181 \\
(h) & 0.884 & 0.022 & 29.17 & 0.280 & 0.883 & 0.029 & 29.47 & 0.154 & 0.910 & 0.018 & 31.55 & 0.213 & 0.953 & 0.014 & 33.57 & 0.180 \\
(i) & 0.943 & 0.011 & 31.99 & 0.181 & 0.879 & 0.023 & 29.40 & 0.159 & 0.910 & 0.018 & 31.53 & 0.213 & 0.951 & 0.015 & 33.41 & 0.183 \\ \bottomrule
\end{tabular}
}
\end{table}

Over a diverse set of detector--matcher backbones and two SfM-prior-based renderers, SARA consistently preserves novel view synthesis quality while enabling a substantially sparser, geometry-aware matching graph. The qualitative and quantitative agreement with exhaustive matching supports the claim that SARA accelerates SfM without degrading the downstream performance of NVS methods that rely on SfM as a prior.

\FloatBarrier
\subsection{Ablation Study}
To validate the design of our two-layer view-graph, we conducted an ablation study to quantify the contribution of each augmentation component: Multi-Scale Loop Consolidation (MSL), Long-Baseline Anchors (LBA), and Weak-View Reinforcement (WVR). We compare our full SARA model against four ablated configurations: Base Only (IWST backbone), w/o MSL, w/o LBA, and w/o WVR. Results are averaged across all scenes and presented in Table \ref{tab:ablation_results}.

\begin{table}[t]
\centering
\caption{Ablation study of augmentation components. We evaluate both removing single components (w/o X) and keeping only single components (Only X) to isolate their contributions.}
\label{tab:ablation_results}
\resizebox{\columnwidth}{!}{
\begin{tabular}{@{}l|cccc@{}}
\toprule
Method & Registered Images & \# 3D Points & Avg. Track Length & Image Pairs \\ \midrule
SARA (Full) & 238 & 48,352 & 9.35 & 589 \\ \midrule
w/o MSL & 238 (0.0\%) & 47,715 (-1.3\%) & 9.27 (-0.8\%) & 589 (0.0\%) \\
w/o LBA & 238 (0.0\%) & 48,181 (-0.4\%) & 9.30 (-0.5\%) & 579 (-1.7\%) \\
w/o WVR & 210 (-11.9\%) & 42,226 (-12.7\%) & 8.92 (-4.6\%) & 495 (-16.1\%) \\ \midrule
Only MSL & 210 (-11.9\%) & 42,033 (-13.1\%) & 8.86 (-5.2\%) & 485 (-17.7\%) \\
Only LBA & 210 (-11.9\%) & 40,766 (-15.7\%) & 9.03 (-3.4\%) & 495 (-16.0\%) \\
Only WVR & 238 (0.0\%) & 47,527 (-1.7\%) & 9.21 (-1.5\%) & 579 (-1.7\%) \\ \midrule
Base Only (IWST) & 210 (-11.9\%) & 40,555 (-16.1\%) & 8.97 (-4.1\%) & 485 (-17.7\%) \\ \bottomrule
\end{tabular}
}
\end{table}

As highlighted in our contributions, Weak-View Reinforcement is critical for achieving complete reconstructions. Disabling WVR causes a major degradation, with an 11.9\% drop in registered images and a 12.7\% drop in 3D point cloud density. The Base Only model shows identical registered image counts, confirming that WVR is solely responsible for recovering these challenging views. This validates our second contribution: targeted edge augmentation for low-confidence views is essential for robustness, preventing reconstruction failure on challenging views that spanning trees alone would miss. In contrast, removing MSL or LBA has minimal impact on completeness, with no loss in registered images. However, their value lies in geometric refinement rather than coverage. Comparing the 'Base Only' model (40,555 points) with 'Only MSL' (42,033 points) and 'Only LBA' (40,766 points) reveals that these components independently enrich reconstruction density even without WVR. While WVR is essential for \textit{completeness} (recovering lost views), MSL and LBA are critical for \textit{quality}. They refine the geometry by closing multi-scale loops and enforcing long-range consistency, contributing to the final leap in point density (from 47,527 in 'Only WVR' to 48,352 in 'Full') and track length. Thus, SARA’s performance relies on the synergy between WVR's structural recovery and MSL/LBA's geometric refinement.

\subsection{Discussion}

Our results show SARA is not just faster, but also more accurate than exhaustive matching. Brute-force matching can actively degrade reconstruction quality by introducing geometric noise from redundant, low-parallax pairs. SARA's sparse, well-conditioned graph provides a cleaner set of constraints for bundle adjustment. The ablation study validates this design, with WVR being essential for completeness and MSL/LBA for geometric quality.

SARA's efficiency relies on the Garbage-In, Garbage-Out principle—it is a high-quality filter, not a corrective tool. SARA correctly rejects low-quality feature matches rather than attempting to salvage them. Additionally, the kNN pre-matching scorer relies on a global descriptor, which could fail in scenes with extreme ambiguity or repetitive structures. Future work could adapt SARA's informativeness-driven selection to real-time SLAM, optimizing keyframe selection and loop closure, or integrate SARA's geometric scoring directly into feature extractor training loops for end-to-end reconstruction-aware feature learning.
\section{Conclusion}
\label{sec:conclusion}

We introduced SARA, a module that shifts pair selection in SfM from visual similarity to geometric information maximization. By scoring pairs on overlap $\times$ parallax, SARA builds a sparse, well-conditioned view-graph that reduces matching complexity from quadratic to quasi-linear while improving reconstruction quality. By filtering geometrically weak pairs, SARA provides cleaner constraints for bundle adjustment, leading to more robust 3D models. As a drop-in module compatible with modern learned detectors, SARA offers a practical path to enhancing the scalability and quality of existing SfM pipelines.

%
%
\paragraph{Acknowledgments.}
This work was supported by the National Research Foundation of Korea (NRF) grant funded by the Korea government (MSIT) (Grant No. RS-2022-NR067080 and RS-2025-05515607).

%
%
%
\bibliographystyle{splncs04}
\bibliography{main}
\end{document}